\documentclass[runningheads]{llncs}

\usepackage[T1]{fontenc}
\def\doi#1{\href{https://doi.org/\detokenize{#1}}{\url{https://doi.org/\detokenize{#1}}}}

\usepackage{graphicx}

\usepackage{listings}
\usepackage{amsfonts}
\usepackage{booktabs}
\usepackage[table]{xcolor}
\usepackage{placeins}
\usepackage{pbox}
\usepackage{amsmath}
\usepackage {xcolor}
\usepackage{multirow}
\usepackage{hyperref}

\begin{document}

\title{Unsupervised Pre-Training on Patient Population Graphs for Patient-Level Predictions}

\titlerunning{Unsupervised Pre-Training on Patient Population Graphs}

\author{Chantal Pellegrini\inst{1} \and
Anees Kazi\inst{1} \and
Nassir Navab\inst{1,2}}

\authorrunning{Chantal Pellegrini, Anees Kazi, Nassir Navab}

\institute{Computer Aided Medical Procedures, Technical University Munich, Germany
\and Computer Aided Medical Procedures, Johns Hopkins University, Baltimore, USA}

\maketitle              

\begin{abstract}

Pre-training has shown success in different areas of machine learning, such as Computer Vision (CV), Natural Language Processing (NLP) and medical imaging. However, it has not been fully explored for clinical data analysis. Even though an immense amount of Electronic Health Record (EHR) data is recorded, data and labels can be scarce if the data is collected in small hospitals or deals with rare diseases. In such scenarios, pre-training on a larger set of EHR data could improve the model performance. In this paper, we apply unsupervised pre-training to heterogeneous, multi-modal EHR data for patient outcome prediction. To model this data, we leverage graph deep learning over population graphs. We first design a network architecture based on graph transformer designed to handle various input feature types occurring in EHR data, like continuous, discrete, and time-series features, allowing better multi-modal data fusion. Further, we design pre-training methods based on masked imputation to pre-train our network before fine-tuning on different end tasks. Pre-training is done in a fully unsupervised fashion, which lays the groundwork for pre-training on large public datasets with different tasks and similar modalities in the future. We test our method on two medical datasets of patient records, TADPOLE and MIMIC-III, including imaging and non-imaging features and different prediction tasks. We find that our proposed graph based pre-training method helps in modeling the data at a population level and further improves performance on the fine tuning tasks in terms of AUC on average by 4.15\% for MIMIC and 7.64\% for TADPOLE.

\keywords{Outcome/Disease Prediction  \and Population Graphs \and Pre-Training.}
\end{abstract}

\section{Introduction}

Enormous amounts of data are collected on a daily basis in hospitals. Nevertheless, labeled data can be scarce, as labeling can be tedious, time-consuming, and expensive. Further, in small hospitals and for rare diseases, only little data is accumulated \cite{mitani2020small}. The ability to leverage the large body of unlabeled medical data to improve in prediction tasks over such small labeled datasets could increase the confidence of AI models for clinical outcome prediction. 
Unsupervised pre-training was shown to be useful to exploit unlabeled data in NLP
\cite{radford2018improving,devlin2018bert}, CV \cite{pathak2016context,bao2021beit} and medical imaging \cite{ouyang2020self,chen2019self}. However, for more complex clinical data, it is not explored enough. Some works study how to pre-train BERT \cite{devlin2018bert} over EHR data of medical diagnostic codes. They pre-train with modified masked language modeling and, in one case, a supervised prediction task. The targeted downstream tasks lie in disease and medication code prediction \cite{shang2019pre,li2020behrt,rasmy2021medbert}. McDermott et al. \cite{mcdermott2021EHRbenchmark} propose a pre-training approach over heterogeneous EHR data, including e.g. continuous lab results. They create a benchmark with several downstream tasks over the eICU \cite{pollard2018eicu} and MIMIC-III \cite{mimic} datasets and present two baseline pre-training methods. Pre-training and fine-tuning are performed over single EHRs from ICU stays with a Gated Recurrent Unit.

On the other hand, population graphs have been leveraged in the recent literature to help analyze patient data using the relationships among patients, leading to clinically semantic modeling of the data. During unsupervised pre-training, graphs allow learning representations based on feature similarities between the patients, which can then help to improve patient-level predictions. Several works successfully apply pre-training to graph data in different domains like molecular graphs and on common graph benchmarks. Proposed pre-training strategies include node level tasks, like attribute reconstruction, graph level tasks like property prediction, or generative tasks such as edge prediction \cite{hu2019strategies,rong2020grover,zhang2020graphbert,hu2020gpt,lu2021learning}.
To the best of our knowledge, no previous work applied pre-training to population graphs.

In this paper, we propose a model capable of pre-training for understanding patient population data. We choose two medical applications in brain imaging and EHR data analysis on the public datasets TADPOLE \cite{tadpole} and MIMIC-III \cite{mimic} for Alzheimer’s disease prediction \cite{parisot2018disease,kazi2019self,cosmo2020latent} and Length-of-Stay prediction \cite{zebin2019deep,wang2020mimicextract,mcdermott2021EHRbenchmark}. The code is available at \href{https://github.com/ChantalMP/Unsupervised-Pre-Training-on-Patient-Population-Graphs-for-Patient-Level-Predictions}{https://github.com/ChantalMP/Unsupervised-Pre-Training-on-Patient-Population-Graphs-for-Patient-Level-Predictions}.\\
\textbf{Contribution:} We develop an unsupervised pre-training method to learn a general understanding of patient population data modeled as graph, providing a solution to limited labeled data. We show significant performance gains through pre-training when fine-tuning with as little as 1\% and up to 100\% labels. Further, we propose a (graph) transformer based model suitable for multi-modal data fusion. It is designed to handle various EHR input types, taking static and time-series data and continuous as well as discrete numerical features into account.

\section{Method}
\begin{figure}[hbt!]
\centering
\includegraphics[width=\textwidth]{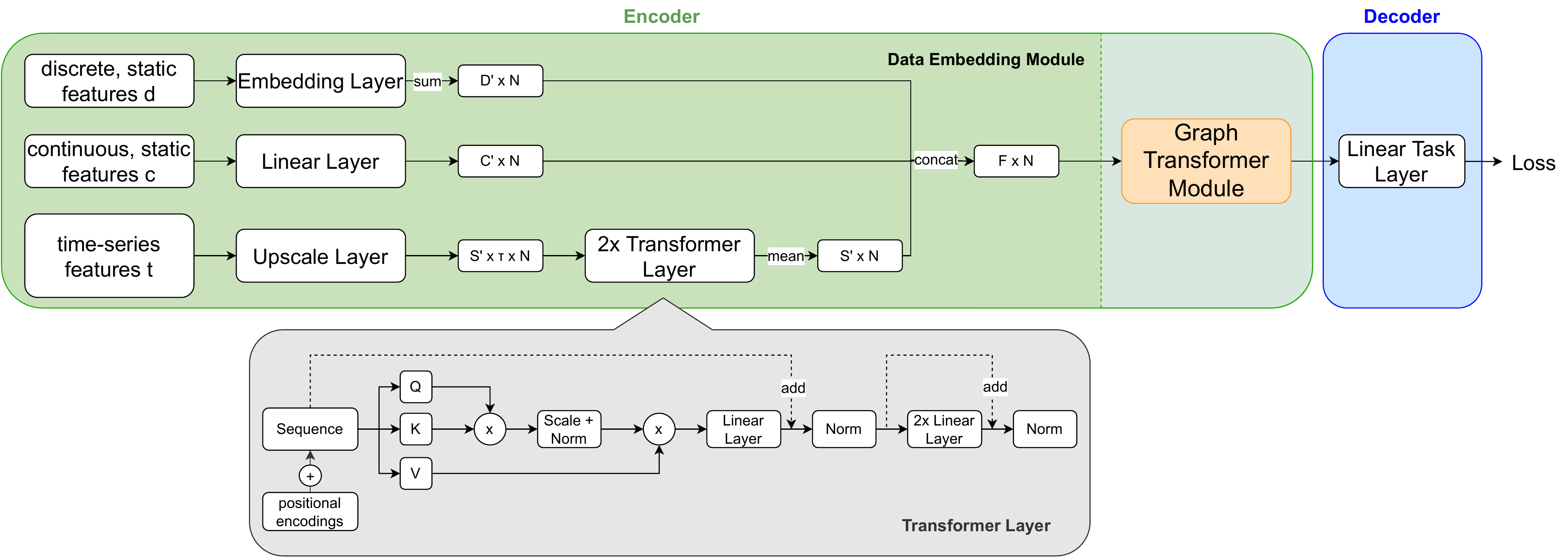}
\caption{Overview of the proposed architecture. All input features are combined into one node embedding, applying transformer layers to enhance the time-series features. The upscale layer for time-series features is a linear layer for continuous and an embedding layer followed by a summation over the feature dimension for discrete features. The resulting graph is processed by several Graphormer layers and a linear task layer.}
\label{arch_overview}
\end{figure}

Let $\mathbf{D}$ be a dataset composed of the EHR data of \textit{N} patients. The $i^{th}$ record is represented by $\mathbf{r_i} \subseteq [\mathbf{d_i},\mathbf{c_i},\mathbf{t_i}]$ with static discrete features $\mathbf{d_i} \in \mathbb{N}^D$ , continuous features $\mathbf{c_i} \in \mathbb{R}^C$, and time-series features  $\mathbf{t_i} \in \mathbb{R}^{S \times \tau }$, where $\tau $ denotes the length of the time-series.
For every downstream task $T$, labels $\mathbf{Y} \in \mathbb{N}^L$ are given for L classes. The task is to predict the classes for the test set patients given all features. Towards this task, we propose to use patient population graphs. 
Unlike in non-graph-based methods, the model can exploit similarities between patients to better understand the EHR data at patient and population level.
Further, unlike conventional graph neural networks, graph transformers allow flexible attention to all nodes, learning which patients are relevant for a task. This would be most apt for learning over population EHR data.
Our pipeline consists of two steps. 1) Unsupervised pre-training and 2) Fine-tuning. Unsupervised pre-training enables understanding of general EHR data and disorder progression by training the model for masked imputation task. This understanding can help learn downstream tasks better despite limited labeled data.\\
\textbf{Graph Construction}
For each node pair with records $r_i$ and $r_j$, we calculate a similarity score $S(r_i, r_j)$ between the node features. We use L2 distance for continuous and absolute matching for discrete features. As graph construction is not our focus, we choose the most conventional method using k-NN selection rule. We set k=5 to avoid having many disconnected components and very densely connected regions (see supplementary material). A detailed description of the graph construction per dataset follows in the experiment section.\\
\textbf{Model Architecture}
Our model consists of an encoder and decoder. The encoder comprises a data embedding module and a graph transformer module explained later. We design the encoder to handle various input data types. The decoder is a simple linear layer capable of capturing the essence of features inclined towards a node-level classification task. Figure \ref{arch_overview} shows an overview of our model architecture.\\
\noindent \textit{Data embedding module:} Following the conventional Graphormer \cite{graphormer}, we process discrete input features by an embedding layer, followed by a summation over the feature dimension, resulting in embedded features $\mathbf{d'_i} \in \mathbb{R}^{D'}$, where $D'$ is the output feature dimension. While Graphormer is limited to static, discrete input features only, we improve upon Graphormer to support also static, continuous input features, which are processed by a linear layer resulting in the embedding vector $\mathbf{c'_i} \in \mathbb{R}^{C'}$. The third branch of our data embedding module handles time-series input features $\mathbf{t_i} \in \mathbb{R}^{S \times \tau}$ with a linear layer, followed by two transformer layers to deal with variable sequence lengths and allow the model to incorporate temporal context. The output is given by $\mathbf{t'_{i,h}} \in \mathbb{R}^{E}$ per time-step h.
The mean of these embeddings forms the final time-series embeddings $\mathbf{t'_i} \in \mathbb{R}^{S'}$. The feature vectors $\mathbf{d'_i}, \mathbf{c'_i}$ and $\mathbf{t'_i}$ are concatenated to form the final node embeddings $n_i \in \mathbb{R}^{F}$, where $F = \sum_{F_k \subset [D', C', S']} F_k$, for each of the \textit{N} nodes.\\
\textit{Graphormer Module:} The backbone of our model comprises multiple graph transformer layers \cite{graphormer}. Graphormer uses attention between all nodes in the graph. To incorporate the graph structure, structural encodings are used, which encode in and out degrees of the nodes, the distance between nodes, and edge features.\\ 
\textbf{Pre-training and Fine-Tuning:} We propose an unsupervised pre-training technique on the same input features as for downstream tasks, but without using labels $\mathbf{Y}$. Instead, we randomly mask a fixed percentage of feature values for every record $\mathbf{r_i}$ and optimize the model to predict these values.  For all methods masking is performed by replacing certain feature values with a fixed value called 'masked token' for discrete features and with zero for continuous features. For time-series features, we further add a binary column per feature to the input vector, that encodes which hours in the time-series are masked. We optimize the model using (binary) cross entropy loss for discrete and mean squared error loss for continuous features. A model for fine-tuning is initialized using the encoder weights learned during pre-training and random weights for the decoder. Then the model is fine-tuned for the task $T$.\\

\section{Experiments and Results}
We use two publicly available medical data sets: TADPOLE \cite{tadpole} and MIMIC-III \cite{mimic}. They differ in size, the targeted prediction task, and the type of input features, allowing comprehensive testing and evaluation of our method.

\subsection{Datasets description:}
\textbf{TADPOLE \cite{tadpole}} contains 564 patients from the Alzheimer’s Disease Neuroimaging Initiative (ADNI). We use twelve features, which the TADPOLE challenge claims are informative. They include discrete cognitive test results, demographics, and continuous features extracted from MR and PET imaging, normalized between zero and one. The task is to classify the patients into the groups Cognitive Normal (CN), Mild Cognitive Impairment (MCI), or Alzheimer’s Disease (AD). We only use data from patients' first visits to avoid leakage of information.\\
\textit{Graph Construction:}
We construct a k-NN graph with k=5, dependent on the mean similarity ($S$) between the features. For the demographics, age, gender and apoe4,
$S_{dem}(r_i, r_j) = \sum \begin{cases}   
1 \textnormal{ if } f_i = f_j  \textnormal{ else } 0 \\  
1 \textnormal{ if } \left|age_{i} - age_{j} \right| \le 2 \textnormal{ else } 0
\end{cases}\div 3
$
where, \textit{f}=(apoe4, gender).
For the cognitive test results $\mathbf{d_i}$ (ordinal features), and $\mathbf{c_i}$ (continuous imaging features), we calculate the respective normalized L2 distances:
$S_{cog}(r_i, r_j) = \frac{\sum_{f \in \mathbf{d_i}}||f_{r_i} - f_{r_j}||}{max(\mathbf{d_i})}
$ and
$S_{img}(r_i, r_j) = sig({\sum_{f \in \mathbf{c_{i}}}||f_{r_i} - \\f_{r_j}||}).
$
The overall similarity $S(r_i, r_j)$ is then given as mean of $S_{dem}$, $S_{cog}$ and $S_{img}$.\\
\textit{Pre-Training Configuration:} During pre-training on TADPOLE, we randomly mask 30\% of the medical features (APOE4, cognitive tests, and imaging features) in each sample. The masking ratio of  30\% was chosen experimentally. \\
\textbf{MIMIC-III \cite{mimic}} is a large EHR dataset of patient records with various static and time-series data collected over the patient’s stay. We use the pre-processed dataset published by McDermott et al. \cite{mcdermott2021EHRbenchmark}. It includes 19.7 K patients that are at least 15 years old and stayed 24 hours or more in the ICU. The features include demographics, measurements from bed-side monitoring and lab tests in hourly granularity (continuous), and binary features stating if different treatments were applied in each hour. In total we have 76 features. We use linear interpolation to impute missing measurements. Fine-tuning is evaluated on  Length-of-Stay (LOS) prediction as defined in \cite{mcdermott2021EHRbenchmark}. The input encompasses the first 24 hours of each patient’s stay, and the goal is to predict if a patient will stay longer than three days or not.\\
\textit{Graph Construction:}
It is computationally infeasible to process a graph containing all patients. Thus, we create sub-graphs with 500 patients each, which fit into memory, each containing train, validation and test patients. We split randomly as we do not want to make assumptions on which types of patients the model should see, but learn this via the attention in the graph transformer. Given the time-series of the measurement features \textit{f}, we form feature descriptors $f_d = (mean(f), std(f), min(f), max(f))$ per patient and feature, where d equals the 56 measurement features. We then compute the average similarity over all features $f_d$ between two patients $r_i$ and $r_j$: $Sim(r_i, r_j) = \frac{\sum_{f \in f_d}||f_{r_i} - f_{r_j}||}{|f_d|}
$
and build a k-NN graph with k=5.\\
\textit{Pre-Training Configuration:}
On MIMIC-III, we perform masking on the time-series features from measurement and treatment data. Pre-training is performed over data from the first 24 hours of the patient’s stay. We compute the loss only over measured values, not over interpolated ones. Masking ratios are chosen experimentally. We compare two types of masking:

\textbf{Feature Masking (FM):} We randomly select 30\% of the features per patient and mask the full 24 hours of the time-series. The model can not see past or future values, only other features and patients, aiming to force an understanding of relations between features and patients to infer masked features.

\textbf{Block-wise Masking (BM):} Instead of the full features, we mask a random block of 6 hours within the 24-hour time-series in 100\% of the features. Here, the model can access past and future values to make a prediction. Thus, it can learn to understand temporal context during pre-training.

\subsection{Experimental Setup}
Given a pre-trained model, we compare the results of fine-tuning it, with training the same but randomly initialized model from scratch.
We manually tuned hyper-parameters per dataset separately for pre-training, from scratch training, and fine-tuning. To simulate scenarios with limited labeled data, we measure the model performance at different label ratios, meaning different amounts of labels (1\%, 5\%, 10\%, 50\%, 100\%) are used for training or fine-tuning. For pre-training always the full training data is used.\\
\textbf{Implementation Details}
All experiments are implemented in PyTorch, performed on a TITAN Xp GPU with 12GB VRAM, and optimized with the Adam optimizer \cite{kingma2014adam}. For cross-validation, pre-training is performed separately per fold. The model comprises four Graphormer layers for TADPOLE and eight for MIMIC-III. For TADPOLE, we pre-train for 6000 epochs with a LR of 1e-5. We train task prediction for 1200 epochs with a polynomial decaying LR (1e-5 to 5e-6) to train from scratch and a LR of 5e-6 for fine-tuning. When fine-tuning with 1\% labels, we reduce the epochs to 200. All results are computed with 10-fold cross-validation. For MIMIC-III, we pre-train for 3000 epochs with a polynomial decaying LR (1e-3 to 1e-4). We train for 1100 epochs with a LR of 1e-4 from scratch, or fine-tune for 600 epochs with a LR of 1e-5. For a fair comparison with the state of the art, results are averaged over six folds, each with an 80-10-10 split into train, validation and test data. The models are selected based on the validation sets, and performance is computed over the test sets.

\subsection{Results}
\textbf{Comparative methods: Table \ref{full_data_results}}
We compare our model to related work without any pre-training. On TADPOLE, we compare to a latent graph learning paper proposed by Cosmo et al. \cite{cosmo2020latent}, which proposes to learn an optimal population graph for the given task. 
Besides, one recent arxiv paper \cite{kazi2021ia} further improves performance on TADPOLE by learning input feature importance. However it is out of context for this work. We achieve comparable accuracy to DGM and outperform in terms of AUC, which is an important metric for imbalanced datasets. For MIMIC-III, we compare our method to the EHR pre-training benchmark of McDermott et al. \cite{mcdermott2021EHRbenchmark}, which uses the same LOS definition and dataset. We significantly outperform the benchmark model. The results show that the proposed architecture is a good fit for the task at hand.
\begin{table}[htb!]
  \caption{Accuracy and AUC of the proposed method compared with DGM on TADPOLE and McDermott et al. \cite{mcdermott2021EHRbenchmark} on MIMIC-III.}
  \centering
    \begin{tabular}{ccc|ccc}
        \toprule
         \multicolumn{3}{c|}{TADPOLE} & \multicolumn{3}{|c}{MIMIC-III} \\
        \midrule
        Model & ACC & AUC & Model & ACC & AUC \\
        \midrule 
        Cosmo \cite{cosmo2020latent} & \textbf{92.91 ± 02.50} & 94.49 ± 03.70 & McDermott \cite{mcdermott2021EHRbenchmark} & - & 71.00 ± 1.00\\
        Proposed & 92.59 ± 3.64 & \textbf{96.96 ± 2.32} & Proposed & 70.29 ± 1.10 & \textbf{76.17 ± 1.02}\\
        \bottomrule
    \end{tabular}
    \label{full_data_results}
\end{table}\\
\textbf{Effect of pre-training: Table \ref{results_table}}
The motivation of this experiment is to investigate the smallest amount of labels required during the fine-tuning of the downstream task. The results emphasize the benefits of our unsupervised pre-training with limited labels.
On TADPOLE the main benefit of pre-training can be seen for settings with limited labels (1\%, 5\%, 10\%), where performance improves significantly. Moreover, AUC continues to improve for all ratios.
For LOS on MIMIC-III, both metrics significantly improve for all label ratios compared to from scratch training. Further, for MIMIC-III we compare two types of masking (BM, FM). We see that feature masking consistently outperforms block-wise masking. The performance improvements achieved through pre-training on MIMIC-III are significantly higher than in the benchmark \cite{mcdermott2021EHRbenchmark}. Moreover, we see improvements until the full dataset size and not only for limited labels. Further the pre-trained models have a lower standard deviation, indicating higher stability.\\
\textbf{Ablation experiments: Table \ref{ablations}}
\begin{table}[htb!]
  \caption{Performance of the proposed model in accuracy and AUC trained from scratch (SC) or fine-tuned after pre-training (FT) for different label ratios. For MIMIC-III we additionally compare the block-wise (BM) and feature masking (FM) to each other.}
    \begin{tabular}{c|c|c|c|c|c|c}
        \toprule
         & & \multicolumn{2}{c|}{TADPOLE} & \multicolumn{3}{|c}{MIMIC-III} \\
        \midrule
        Size & Metric & SC & FT & SC & FT: BM & FT: FM \\
        \midrule
        1\% & ACC & 59.42 ± 8.40 & \textbf{78.89 ± 2.45} & 59.86 ± 2.11 & \color{blue}63.22 ± 2.39 & \textbf{65.25 ± 1.09}\\
        & AUC & 68.72 ± 12.74 & \textbf{93.49 ± 2.07} & 62.98 ± 2.55 & \color{blue}68.07 ± 1.80 & \textbf{69.90 ± 1.26}\\
        \midrule
        5\% & ACC & 78.23 ± 6.83 & \textbf{83.37 ± 6.29} & 64.79 ± 1.16 & \color{blue}66.82 ± 0.89 & \textbf{68.66 ± 0.73}\\
        & AUC & 87.23 ± 4.91 & \textbf{94.99 ± 2.55} & 68.85 ± 1.53 & \color{blue}72.27 ± 1.19 & \textbf{73.97 ± 1.28}\\
        \midrule
        10\% & ACC & 87.00 ± 4.86 & \textbf{87.71 ± 4.65} & 64.72 ± 0.45 & \color{blue}67.71 ± 0.69 & \textbf{69.42 ± 1.23}\\
        & AUC & 92.03 ± 3.39 & \textbf{95.96 ± 2.51} & 68.97 ± 0.66 & \color{blue}73.55 ± 0.60 & \textbf{75.09 ± 1.29}\\
        \midrule
        50\% & ACC & \textbf{92.41 ± 3.69} & 91.52 ± 3.76 & 67.41 ± 1.31 & \color{blue}69.98 ± 0.69 & \textbf{70.85 ± 0.92}\\
        & AUC & 96.06 ± 2.48 & \textbf{97.23 ± 1.94} & 72.53 ± 1.08 & \color{blue}76.02 ± 0.87 & \textbf{76.86 ± 1.47}\\
        \midrule
        100\% & ACC & \textbf{92.59 ± 3.64} & 92.24 ± 3.47 & 70.29 ± 1.10 & \color{blue}70.73 ± 0.70 & \textbf{71.44 ± 1.25}\\
        & AUC & 96.96 ± 2.23 & \textbf{97.52 ± 1.67} & 76.17 ± 1.02 & \color{blue}76.20 ± 0.54 & \textbf{77.78 ± 1.31}\\
        \bottomrule
    \end{tabular}
    \label{results_table}
\end{table}
We perform several ablation studies to evaluate different parts of our proposed model on pre- and task training.\\
\textit{Effect of Graphormer:} We replace the Graphormer module with a simple linear or GCN layer and train the model from scratch on the full dataset (Table \ref{ablations} a)). We see a clear benefit from using Graphormer compared to the linear model and GCN. For TADPOLE, the linear model reaches slightly better performance in terms of AUC as TADPOLE is a relatively small and easy dataset. The effect of the node level attention mechanism to all nodes given by Graphormer is clearly visible when compared to GCN. Further, we perform pre-training followed by fine-tuning for the linear model (Table \ref{ablations} b)). Our proposed unsupervised pre-training method proves to be beneficial also for the linear model, but the effects are less as for our proposed architecture. Table \ref{ablations} c) shows masked imputation performance during pre-training, measured by RMSE for continuous (imaging/ measurements) and accuracy or F1 for discrete features (apoe4+cognitive tests/treatments). Here the proposed model outperforms the linear model, explaining why pre-training has a greater effect for it. In summary we see a positive effect of using Graphormer over the linear model for solving the pre-training task and improving fine-tuning performance.\\
\textit{Effect of Transformer:}
For MIMIC-III, Transformer is inserted in the encoder to deal with time series data. To test the transformer layers, we remove this component and train the model from scratch on the full dataset, resulting in a reduction of accuracy from 70.29 to 69.39\% and AUC from 76.17 to 75.03\%. This shows that the transformer layers are helpful for processing time-series inputs. The model needs to predict time-dependent outputs for pre-training on MIMIC-III, for which the transformer layers are important, as they can understand the temporal context. To investigate the effect of transformer during pre-training, we remove the transformer layer and replace the Graphormer module with an linear layer. We observe a reduction in the performance by 0.45\% for ACC and 3.03\% in AUC through pre-training. Accordingly, removing the transformer layer results in a 0.049 larger RMSE and a 3.9\% lower F1 score in pre-training.
\begin{table}[hbt!]
  \caption{Ablations to test Graphormer module by replacing it with a linear/GCN layer, a) downstream task performance trained from scratch b) results of fine-tuning (FT) on limited labels (TADPOLE 1\%, MIMIC-III 10\%), compared to training from scratch (SC) c) pre-training task performance, multi-class accuracy for cognitive tests uses feature-dependent error margins in which predictions are considered correct. The small number of imaging features might cause the low std of 0.006/0.008.}
  \centering 
    \begin{tabular}{p{0.4cm}|p{2cm}p{2.3cm}p{2.3cm}|p{2.3cm}p{2.3cm}}
        \toprule
        & \multicolumn{3}{c|}{TADPOLE} & \multicolumn{2}{|c}{MIMIC-III} \\
        \midrule
        &Model & ACC & AUC & ACC & AUC \\
        \midrule
        &Linear & 91.14 ± 02.62 & \textbf{97.77 ± 01.59} & 67.25 ± 01.11 & 72.69 ± 00.97\\
        \multirow{3}{*}{a} & GCN & 74.27 ± 06.41 & 89.89 ± 04.12 & 68.74 ± 01.50 & 72.64 ± 01.10\\
        &Proposed & \textbf{92.59 ± 03.64} & 96.96 ± 02.23 & \textbf{70.29 ± 01.10} & \textbf{76.17 ± 01.02}\\
        \midrule
        &Linear SC & 54.20 ± 08.74 & 70.41 ± 11.41 & 63.78 ± 00.74 & 67.72 ± 00.68\\
        \multirow{4}{*}{b} & Linear FT & \textbf{71.27 ± 09.76} & \textbf{89.25 ± 06.53} & \textbf{64.71 ± 00.84} & \textbf{67.94 ± 01.20} \\
        &Proposed SC & 59.42 ± 08.40 & 68.72 ± 12.74 &  64.72 ± 00.45 & 68.97 ± 00.66\\
        &Proposed FT & \textbf{78.89 ± 02.45} & \textbf{93.49 ± 02.07} & \textbf{69.42 ± 01.23} & \textbf{75.09 ± 01.29}\\
        \midrule
        && \parbox{2.4cm}{\centering RMSE} & \parbox{2.4cm}{\centering ACC} & \parbox{2.4cm}{\centering RMSE} & \parbox{2.4cm}{\centering F1} \\
        \midrule
      \multirow{2}{*}{c} & Linear & 00.15 ± 0.008 & 62.58 ± 04.87 & 00.79 ± 0.023 & 81.49 ± 00.35\\
       & Proposed & \textbf{0.14 ± 0.006} & \textbf{63.23 ± 04.25} & \textbf{0.78 ± 0.011} & \textbf{81.58 ± 00.41}\\
        \bottomrule
    \end{tabular}
    \label{ablations}
\end{table}

\FloatBarrier
\section{Conclusion}
In this paper, we present an unsupervised pre-training method based on masked imputation, significantly improving prediction results. We propose a graph transformer based architecture for learning on population graphs built from heterogeneous EHR data. We show the superiority of our pipeline in both pre-training and various prediction tasks for two datasets, TADPOLE and MIMIC-III. Pre-training helps for all dataset sizes but especially in scenarios where only a limited amount of labeled data is used for fine-tuning.
Our pre-training method is unsupervised and therefore independent from the end task, and further it is well suited for transfer learning. This work opens the path for the community to deals with small dataset specially with limited labels.
%

\bibliographystyle{splncs04}
\bibliography{refs}

\end{document}